\documentclass{article}

    \PassOptionsToPackage{numbers, compress}{natbib}
    \usepackage[preprint]{neurips_2024}

\usepackage[utf8]{inputenc} % allow utf-8 input
\usepackage[T1]{fontenc}    % use 8-bit T1 fonts
\usepackage{url}            % simple URL typesetting
\usepackage{booktabs}       % professional-quality tables
\usepackage{amsfonts}       % blackboard math symbols
\usepackage{amsmath}        % Add this line to include the amsmath package
\usepackage{amssymb}
\usepackage{nicefrac}       % compact symbols for 1/2, etc.
\usepackage{microtype}      % microtypography
\usepackage{xcolor}         % colors
\usepackage{graphicx}

\usepackage{caption}
\usepackage{subcaption}
\usepackage{pifont}
\usepackage{csquotes}

\newcommand{\cmark}{\ding{51}} % Check mark
\newcommand{\xmark}{\ding{55}} % Cross mark

\usepackage[colorlinks]{hyperref} 
\usepackage[capitalize]{cleveref}

\title{ChatSchema: A pipeline of extracting structured information with Large Multimodal Models based on schema}

\author{%
  Fei Wang\thanks{These authors contributed equally to this study.} \\
   Advanced Institute of \\
   Information Technology,\\
  Peking University\\
  Hangzhou 311215, China  \\
  \texttt{fei.comm@icloud.com} \\
  \And
  Yuewen Zheng$^*$ \\
Advanced Institute of \\
Information Technology,\\
  Peking University\\
  Hangzhou 311215, China \\
  \texttt{zhengyuewen27@dingtalk.com} \\
    \And
  Qin Li \\
Advanced Institute of \\
Information Technology,\\
  Peking University\\
  Hangzhou 311215, China \\
  \texttt{blacknepia@dingtalk.com} \\
    \AND
  Jingyi Wu \\
Advanced Institute of \\
Information Technology,\\
  Peking University\\
  Hangzhou 311215, China \\
  \texttt{joywu@pku.edu.cn} \\
  \And
  Pengfei Li \thanks{Corresponding to Pengfei Li, pfli@bjmu.edu.cn} \\
  Advanced Institute of \\
  Information Technology, \\
  Peking University \\
  Hangzhou 311215, China  \\
  \texttt{pfli@bjmu.edu.cn} \\
  \And
  Luxia Zhang \\
  National Institute of \\
  Health Data Science, \\
  Peking University \\
  Beijing 100191, China \\
  \texttt{zhanglx@bjmu.edu.cn} \\
}

\begin{document}

\maketitle

\begin{abstract}
  Objective: This study introduces ChatSchema, an effective method for extracting and structuring information from unstructured data in medical paper reports using a combination of Large Multimodal Models (LMMs) and Optical Character Recognition (OCR) based on the schema. By integrating predefined schema, we intend to enable LMMs to directly extract and standardize information according to the schema specifications, facilitating further data entry.
  Method: Our approach involves a two-stage process, including classification and extraction for categorizing report scenarios and structuring information. We established and annotated a dataset to verify the effectiveness of ChatSchema, and evaluated key extraction using precision, recall, F1-score, and accuracy metrics. Based on key extraction, we further assessed value extraction. We conducted ablation studies on two LMMs to illustrate the improvement of structured information extraction with different input modals and methods.
  Result: We analyzed 100 medical reports from Peking University First Hospital and established a ground truth dataset with 2,945 key-value pairs. We evaluated ChatSchema using GPT-4o and Gemini 1.5 Pro and found a higher overall performance of GPT-4o. The results are as follows: For the result of key extraction, key-precision was 98.6\%, key-recall was 98.5\%, key-F1-score was 98.6\%. For the result of value extraction based on correct key extraction, the overall accuracy was 97.2\%, precision was 95.8\%, recall was 95.8\%, and F1-score was 95.8\%. An ablation study demonstrated that ChatSchema achieved significantly higher overall accuracy and overall F1-score of key-value extraction, compared to the Baseline, with increases of 26.9\% overall accuracy and 27.4\% overall F1-score, respectively.
  Conclusion: The proposed method significantly improved the extracting and information structuring from medical reports, enabling better identification and standardization of keys and values. Our research indicates that using LMMs with prompt engineering and the OCR text has great potential for enhancing the data entry processing of medical documents.
\end{abstract}

\section{Introduction}

Medical reports are prevalent and require extensive manual work before data entry. This work includes converting data types, performing unit conversions, mapping dictionary-type data, and normalizing fields~\cite{RN37, RN36, RN34}. Recent advancements in artificial intelligence have introduced promising solutions to these challenges, particularly with the emergence of Large Multimodal Models (LMMs). LMMs integrate the capabilities of both Large Language Models (LLMs) and Large Vision Models (LVMs), allowing them to process and reason with multimodal information, including text and images~\cite{RN18}. This integration is particularly beneficial for the extraction of information from medical reports, as it enables the models to handle diverse data formats. For example, Sarmah et al. combines LMMs with OCR technology, making it suitable for the automated recognition and processing of various general domain documents~\cite{RN12}. However, due to the lack of integration with schema specifications, it cannot quickly extract results and standardize values for data entry.

LMMs have demonstrated significant potential in various applications by leveraging their ability to follow instructions, perform in-context learning, and reason through chains of thought~\cite{RN19, RN40, RN21}. Their application in medical information extraction is promising, as these models can use multimodal instruction tuning to follow complex instructions and generate accurate outputs from multimodal data sources. The new training paradigms and billion-scale parameters of LMMs provide them with unique capabilities, such as writing code based on images and performing OCR-free math reasoning, which is rare in traditional methods~\cite{RN15, RN13, RN5, RN6}.

A crucial aspect of optimizing the performance of LMMs is prompt engineering. Prompt engineering involves designing and refining input prompts to guide the model towards generating the most relevant and accurate outputs~\cite{RN22, RN24}. This technique leverages the inherent capabilities of LMMs to understand and respond to complex instructions, thereby enhancing their ability to perform specific tasks. The potential of prompt engineering extends across various applications. In natural language processing tasks, it has been shown to improve the performance of models in tasks such as text summarization, translation, and question-answering. In the context of medical data, prompt engineering enables models to generate contextually appropriate and precise responses, which is critical for accurate information extraction and standardization~\cite{RN26, RN27, RN28, RN25}.

We apply prompt engineering to LMMs, presenting an efficient approach for generating structured information from unstructured medical records. By carefully designing prompts that align with a predefined schema, we can direct the LMMs to accurately extract and standardize relevant medical entities and values. This method leverages the model's ability to follow detailed instructions, ensuring that the extracted information is not only accurate but also formatted according to the necessary specifications for data entry. By providing examples and clear extraction criteria, the use of prompt engineering allows the LMMs to handle the unstructured data, including the varying terminologies and format conversion. This approach enhances the overall performance of the information extraction process, making it more robust and scalable for diverse medical datasets. Through this method, we can significantly reduce the time and effort required for manual data entry and processing, facilitating more efficient and reliable use of medical reports in clinical research.

\section{Method}
\subsection{Overview}
\begin{figure}
  \centering
  \includegraphics[width=0.8\linewidth]{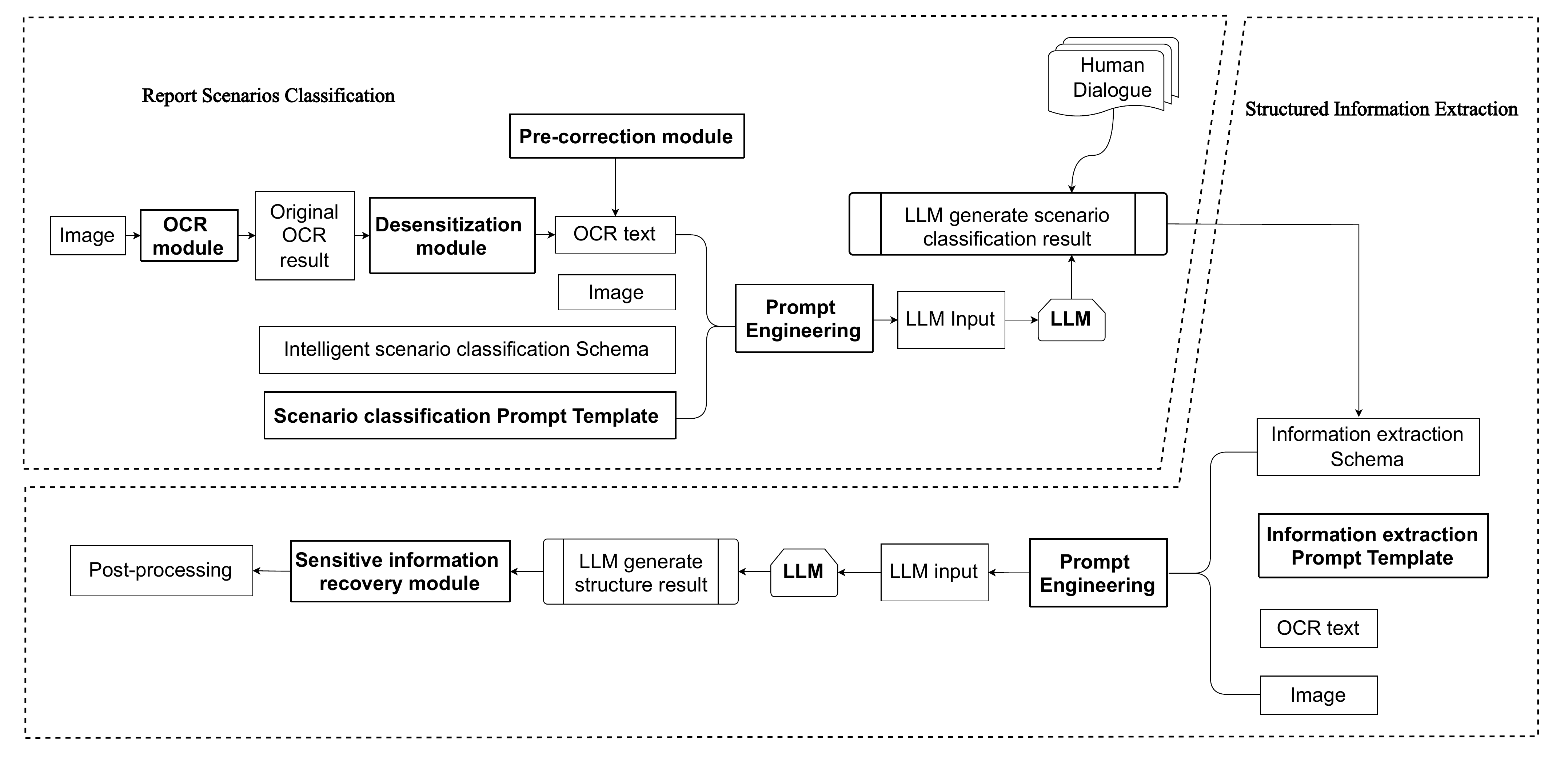}
  \caption{Overview of ChatSchema.}
  \label{fig:overview}
\end{figure}

The architecture of ChatSchema is structured into two principal stages: classification and extraction. The classification stage categorizes the medical report scenarios, while the extraction stage gains structured information from each specific type. We can input both the OCR text and the image into a LMMs. Before feeding the data into the LMMs, we design distinct prompts with the corresponding schema for classification and extraction. An overview architecture can be found in \cref{fig:overview}, with the input and output of each stage showcased.

\subsubsection{Report scenarios classification}
\begin{figure}
  \centering
  \includegraphics[width=0.8\linewidth]{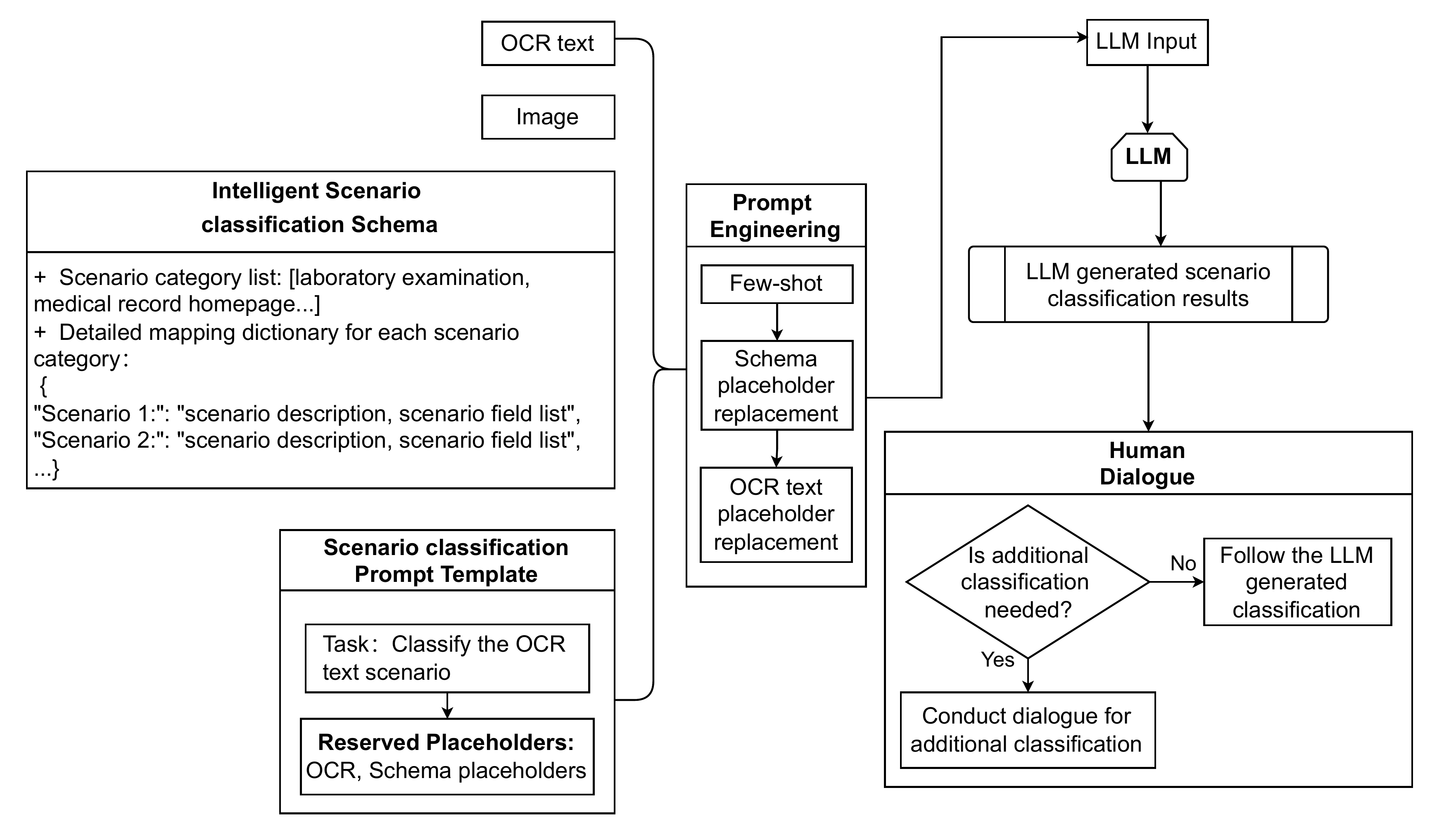}
  \caption{Details of report scenarios classification stage.}
  \label{fig:classification}
\end{figure}

The report scenarios classification stage consists of several key components: the OCR module, pre-correction module, desensitization module, and prompt engineering. These components ensure that the OCR text is accurately recognized, corrected, and classified. A detail of report scenarios classification stage can be found in \cref{fig:classification}.

The OCR module preprocesses the input images with techniques such as binarization, noise reduction, and contrast adjustment. Then the text detection is applied to the preprocessed images, resulting in bounding boxes of all text regions. These bboxes are passed to text recognition algorithms to extract the text within each bbox. The recognized text segments are stitched together based on their bbox positions to reconstruct the original text lines, yielding a raw OCR output.

The raw output from OCR frequently contains inaccuracies in text recognition and issues with bounding box misalignment, which can impair the subsequent stages of information extraction. To address these issues, we utilize the contextual understanding capabilities of LMMs in the pre-correction module. The raw OCR text is formatted into a prompt template, within which potential errors including positional errors and visually similar characters such as \enquote{l} and \enquote{1}, are specifically highlighted for attention. This template is then sent to the LMMs and generates corrected text. The template was sent to LMMs and the corrected text was generated by LMMs, thereby reducing the errors propagated to the extracted information.

\begin{figure}
  \centering
  \includegraphics[width=0.8\linewidth]{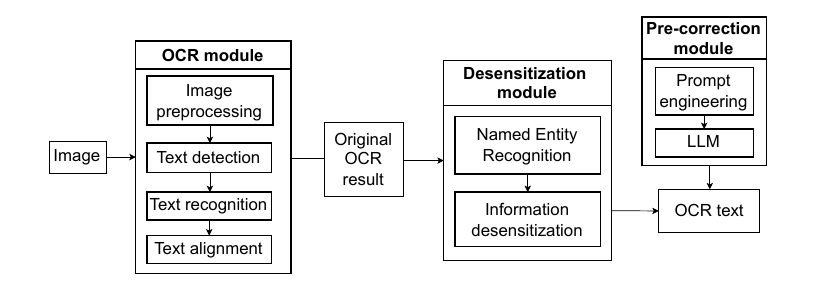}
  \caption{Detail of image preprocessing flow.}
  \label{fig:preprocessing}
\end{figure}

Before inputting the corrected OCR text to scenarios classification, sensitive information entities should be masked to safeguard patient privacy. We employ a Named Entity Recognition (NER) model to identify and anonymize entities such as patient names and identification numbers within the text. Concurrently, a mapping table is maintained to record the reflection between the original sensitive information and its desensitized versions. The details of image preprocessing flow can be found in \cref{fig:preprocessing}.

The desensitized OCR text, paired with the corresponding image and scenario classification schema, is integrated into the scenario classification prompt template, which plays a pivotal role in the prompt engineering phase. The schema comprises a structured list of categories from a project-related schema, along with descriptions of their distinguishing characteristics, facilitating the LMMs in distinguishing between various scenarios. The prompt template defines the task instructions, specifying the purpose of the task and the expected output format. 	The prompt template includes placeholders for the schema and the desensitized OCR text, which are then replaced with the actual data to form the input for the LMMs. This input is subsequently submitted to the LMMs to perform the classification of scenarios. Generally, the medical report can be classified into 1--3 types. For example, one report may include a complete blood count, urinalysis, and comprehensive metabolic panel.

To enhance the accuracy of the LMMs's classification, we apply few-shot learning techniques within the task instructions of the prompt template. Specifically, we provide the LMMs with a few annotated examples. For example, the template may include a conditional directive that, when the OCR text includes medical terms like \enquote{Total Iron Binding Capacity (TIBC)} and \enquote{Serum Ferritin (SF)}, the document is likely a \enquote{Five Iron Profile} medical report. By integrating such directives into the prompt template, the model could recognize new scenario types and classify documents accordingly, even when only a small number of examples are provided.

In the LMMs classification process, the prompt template could be dynamically updated through human dialogue if the existing categories do not comprehensively cover the scenario types. Users can define new scenarios and their corresponding characterizations. The updated scenarios are integrated into the prompt template, expanding the LMMs's capacity to classify the newly defined types.

\subsubsection{Structured information extraction}
\begin{figure}
  \centering
  \includegraphics[width=0.8\linewidth]{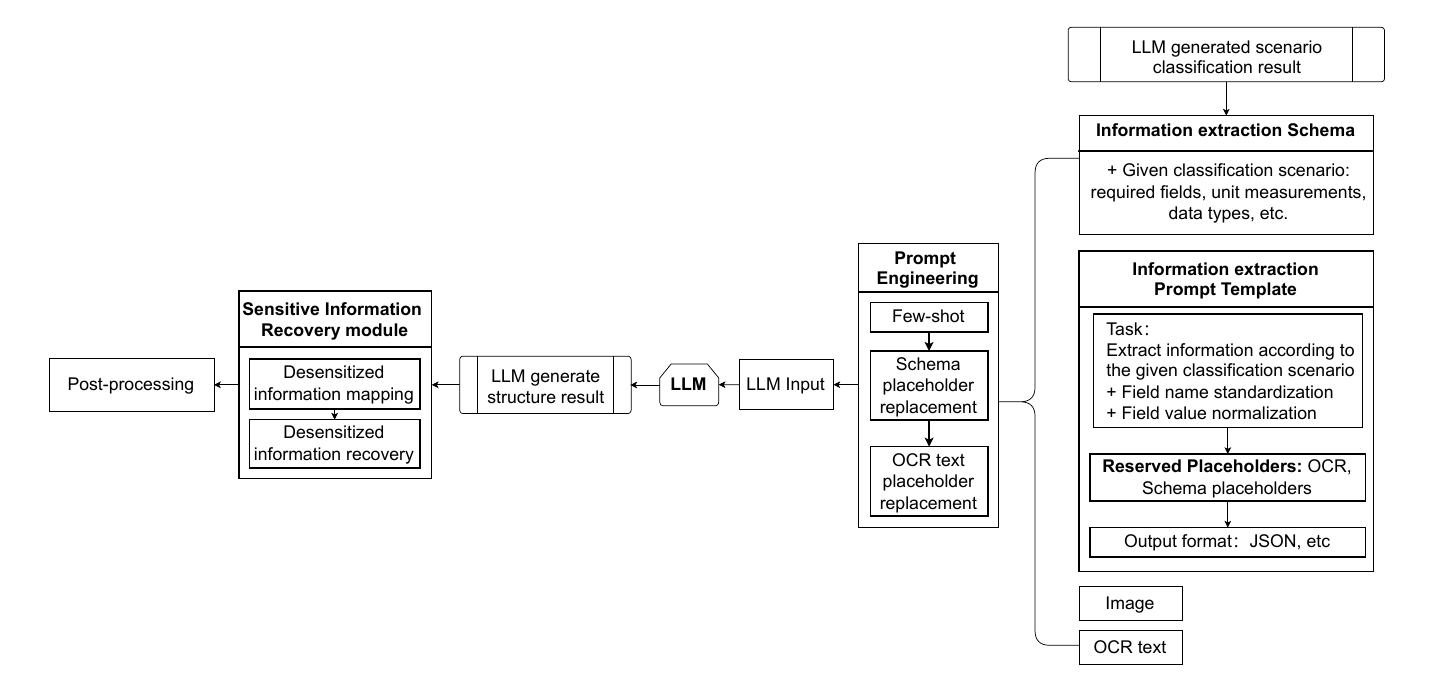}
  \caption{Details of report information extraction stage.}
  \label{fig:structure}
\end{figure}

The structured information extraction stage transforms OCR text into a well-structured, database-ready format based on a predefined schema. This stage comprises prompt engineering and an information recovery module. An overview of the structured information extraction stage can be found in \cref{fig:structure}.

To make OCR text ready for data entry, analysis, and further processing, we first define the schema for information extraction. The schema includes each category and the fields it contains, specifying field names, value types, and units. This schema is a criteria to standardize the key-value pairs according to the specified field requirements and structures from the OCR text.

The LMMs process the implicit key-value pairs in the OCR text, performing field name standardization, unit conversion, value mapping, and data type transformation. The prompt template specifies the task's objective, which is to extract the specified fields from the OCR text based on the given schema. It also includes placeholders for the schema and OCR text and is replaced with actual data to generate the LMMs input. Additionally, the prompt template has instructions for standardizing field names, such as changing \enquote{Sed rate} to \enquote{Erythrocyte Sedimentation Rate (ESR)}, converting units for numerical values -- for example, converting \enquote{3m} to \enquote{300 cm} if the context is meters to centimeters -- mapping field values to predefined options, such as mapping \enquote{+} to \enquote{positive}, and changing data types, such as converting integer values to string format.

Finally, the desensitization module restores the masked sensitive information based on the original mappings, making the final data entered into the authentic.

\subsection{Experiments}
We evaluated the performance of ChatSchema on LMMs of Gemini 1.5 Pro and GPT-4o. To analyze the prediction results of ChatSchema, we divided the output cases into four specific situations: Key-Value match, represented as Corrects, which is both the key and value are rightly detected; Value mismatch, that is, the key is correctly classified, but the value associated with the key is incorrect, which refers to key-True Positives(key-TP); Key not found, this situation arises when a key should be detected but not be identified by the model, which refers to False Negatives(key-FN); An extra key, in this case, the model detects an additional key but it is not present in the annotation and is considered as False Positives(key-FP). We calculate evaluation metrics including key-precision, key-recall, and key-F1-score. Additionally, we determined the overall accuracy, precision, recall, and F1-score to comprehensively assess ChatSchema's performance.

To explore the usefulness of different input modals, we conducted an ablation study using the following three combinations on ChatSchema:
\begin{itemize}
  \item Image only: LMMs were provided with only images of the medical reports.
  \item Text only: LMMs were provided with only OCR-extracted text from the medical reports.
  \item Both (Images and Text): LMMs were provided with both images and OCR-extracted text from the medical reports.
\end{itemize}
We evaluated the Corrects, key-TP, key-FP, and key-FN for all three groups on both LMMs (Gemini 1.5 Pro and GPT-4o) to compare their performance. We calculated their corresponding key-precision, key-recall, and key-F1-score, as well as the overall accuracy, precision, recall, and F1-score.

Furthermore, to demonstrate the effectiveness of ChatSchema, we set Baseline which involved providing only the schema and task instructions, without performing our first classification and then extraction method. To compare the differences between ChatSchema and the Baseline, we conducted experiments under the same input modalities and LMMs. We also evaluated their key-precision, key-recall, and key-F1-score, along with the overall accuracy, precision, recall, and F1-score.

The details of these metrics are as followings:

Key-precision: This is the ratio of correctly detected keys (True Positives, key-TP) to all detected keys (including those incorrectly detected, i.e., False Positives, key-FP). Formula:
\[
  P_\text{key} = \frac {\text{TP}_\text{key}}{\text{TP}_\text{key} + \text{FP}_\text{key}}
\]
where \(\text{TP}_\text{key}\) represents key True Positives, \(\text{FP}_\text{key}\) represents key False Positives.

Key-recall: This is the ratio of correctly detected keys (True Positives, key-TP) to all actual keys that should be detected (including those not detected, i.e., False Negatives, key-FN). Formula:
\[
  R_\text{key} = \frac {\text{TP}_\text{key}}{\text{TP}_\text{key} + \text{FN}_\text{key}}
\]

where \(\text{FN}_\text{key}\) represents key False Positives.

Key-F1-Score: This is the harmonic mean of Key-precision and Key-recall, reflecting the model's balanced performance in these two aspects.
Formula:
\[
  F1_\text{key} = 2\times \frac{ P_\text{key} \times R_\text{key}}{ P_\text{key} + R_\text{key}}
\]

Overall Accuracy: This is the ratio of correctly predicted key-value pairs to the total number of detected keys. Formula:
\[
  Acc = \frac{N}{\text{TP}_\text{key}}
\]

where \(N\) represents the total number of correctly predicted key-value pairs, that is Corrects.

Precision: This is the multiple of key precision and overall Accuracy. Formula:
\[
  P = P_\text{key} \times Acc
\]

Recall: This is the multiple of key recall and overall Accuracy. Formula:
\[
  R = R_\text{key} \times Acc
\]

F1-Score: This is the harmonic mean of precision and recall, reflecting the model's balanced performance in these two aspects.
Formula:
\[
  F1 = 2\times \frac{P \times R}{P + R}
\]

\section{Results}
\subsection{Dataset description}
This study included 100 Peking University First Hospital medical reports. We manually annotated these reports in a JSON format and established a ground truth dataset with 2,945 key-value pairs. The schema has 13 distinct classes, such as iron panels, complete blood counts, bone metabolism panels, etc. Under each category, the schema defined their corresponding detailed items. The total number of detailed items in all categories is 137. Except for the items that can reflect a patient's health in a specific aspect, the schema also contained 18 variables of general information like the patient’s name, gender, department, etc. The measurement units are across 30 different formats, including \enquote{U/g}, \enquote{U/L}, and \enquote{HP}. The data types for the items are diverse, spanning 5 kinds: datetime, integer, string, float, and dictionary. This research was approved by the Ethics Committee of Peking University First Hospital (No. 2019 [146]), and written informed consent was obtained from all participants. The annotated dataset is presented in Appendix A.

\subsection{Experiments result}

\begin{table}
  \caption{The ablation study results on GPT-4o}
  \label{tab:ablation_gpt4o}
  \centering
  \begin{tabular}{@{}ccccccccc@{}}
    \toprule
    Text   & Image  & Key-P         & Key-R         & Key-F1        & Acc           & P             & R             & F1            \\
    \midrule
    \xmark & \cmark & 98.5          & 98.0          & 98.2          & 80.7          & 79.5          & 79.1          & 79.3          \\
    \cmark & \xmark & \textbf{98.7} & \textbf{99.1} & \textbf{98.9} & 97.0          & 95.7          & \textbf{96.1} & \textbf{95.9} \\
    \cmark & \cmark & 98.6          & 98.5          & 98.6          & \textbf{97.2} & \textbf{95.8} & 95.8          & 95.8          \\
    \bottomrule
  \end{tabular}
\end{table}

\begin{table}
  \caption{The ablation study results of Gemini 1.5 Pro}
  \label{tab:ablation_gemini}
  \centering
  \begin{tabular}{@{}ccccccccc@{}}
    \toprule
    Text   & Image  & Key-P         & Key-R         & Key-F1        & Acc           & P             & R             & F1            \\
    \midrule
    \xmark & \cmark & 96.1          & 84.6          & 90.0          & 82.2          & 79.0          & 69.5          & 74.0          \\
    \cmark & \xmark & \textbf{98.3} & 98.7          & 98.5          & \textbf{95.2} & \textbf{93.5} & \textbf{94.0} & \textbf{93.7} \\
    \cmark & \cmark & \textbf{98.3} & \textbf{98.8} & \textbf{98.6} & 94.8          & 93.2          & 93.7          & 93.5          \\
    \bottomrule
  \end{tabular}
\end{table}

The ablation study results of various modals on GPT-4o are shown in \cref{tab:ablation_gpt4o} and Gemini 1.5 Pro are shown in \cref{tab:ablation_gemini} respectively. Using the GPT-4o result as illustrated, when using Image only, the model achieved a relatively low overall accuracy (Acc) of 80.7\%, with a precision (P) of 79.5\%, recall (R) of 79.1\%, and F1-score of 79.3\%. When using Text only, ChatSchema's performance improved significantly, achieving the best overall F1-score of 95.9\%. Employing both images and texts resulted in a performance comparable to text-only input. The model achieved the best overall accuracy of 97.2\%.

These results indicate that using text-only input yields a significant improvement over image-only input. However, combining both images and text does not provide a substantial advantage over using text-only input alone, suggesting that the additional information from images has a minimal impact on performance when text is already available. Therefore, both text-only and text+image approaches are comparable in their effectiveness.

\begin{figure}
  \centering
  \begin{subfigure}[b]{0.3\textwidth}
    \centering
    \includegraphics[width=\textwidth]{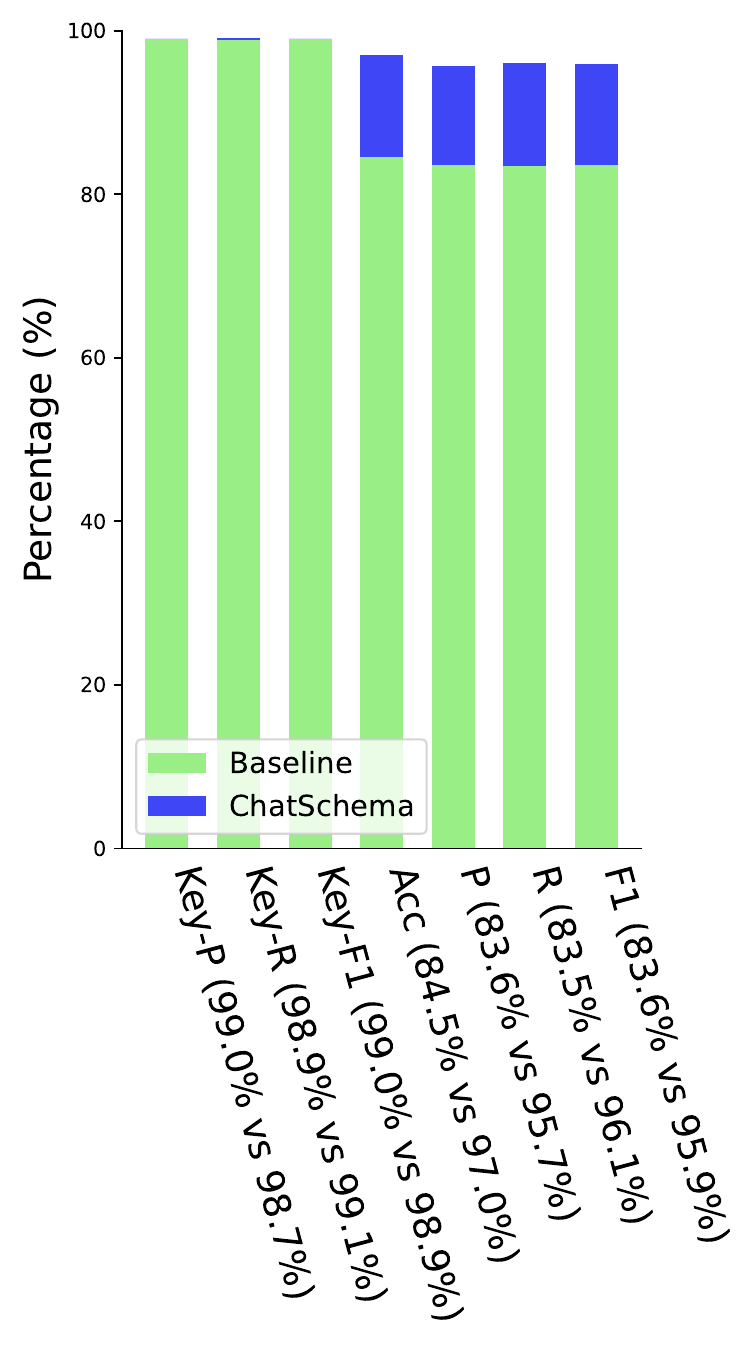}
    \caption{$Image$}
    \label{fig:gpt_image}
  \end{subfigure}
  \hfill
  \begin{subfigure}[b]{0.3\textwidth}
    \centering
    \includegraphics[width=\textwidth]{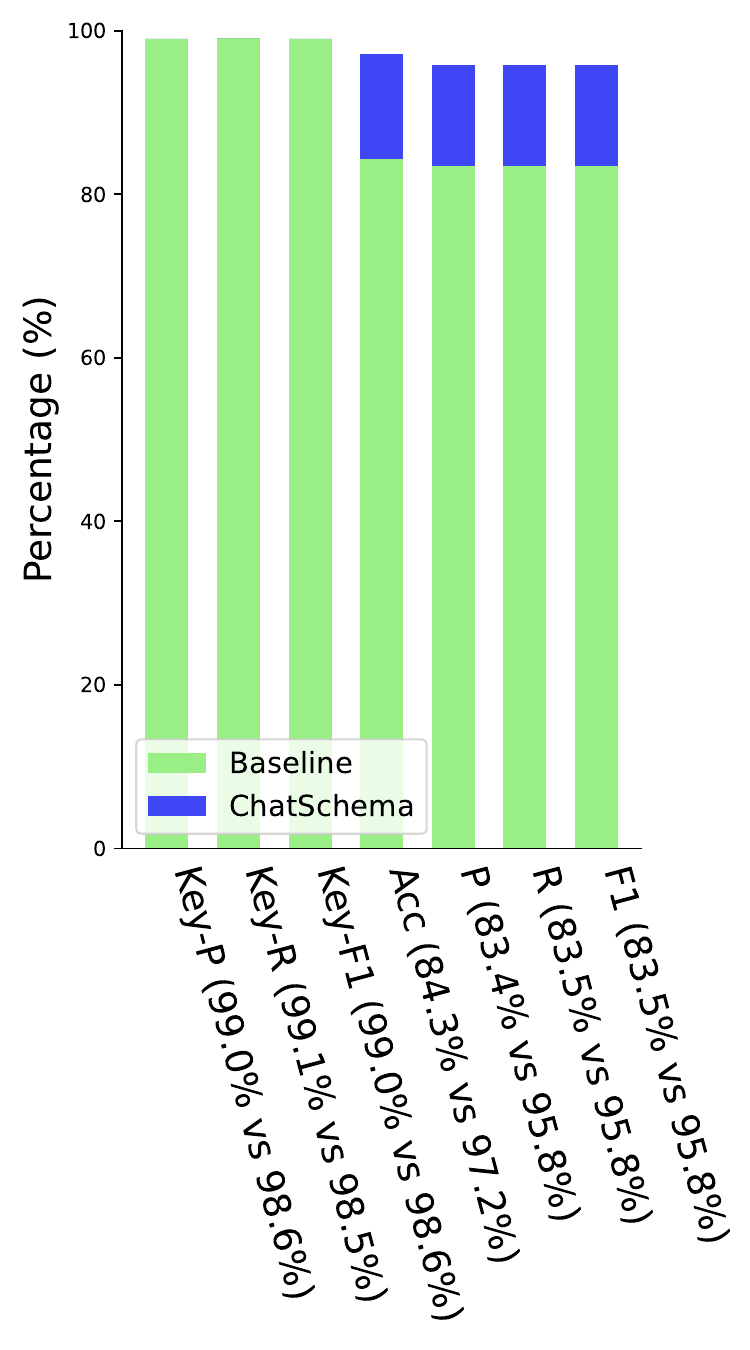}
    \caption{$Image + Text$}
    \label{fig:gpt_mixture}
  \end{subfigure}
  \hfill
  \begin{subfigure}[b]{0.3\textwidth}
    \centering
    \includegraphics[width=\textwidth]{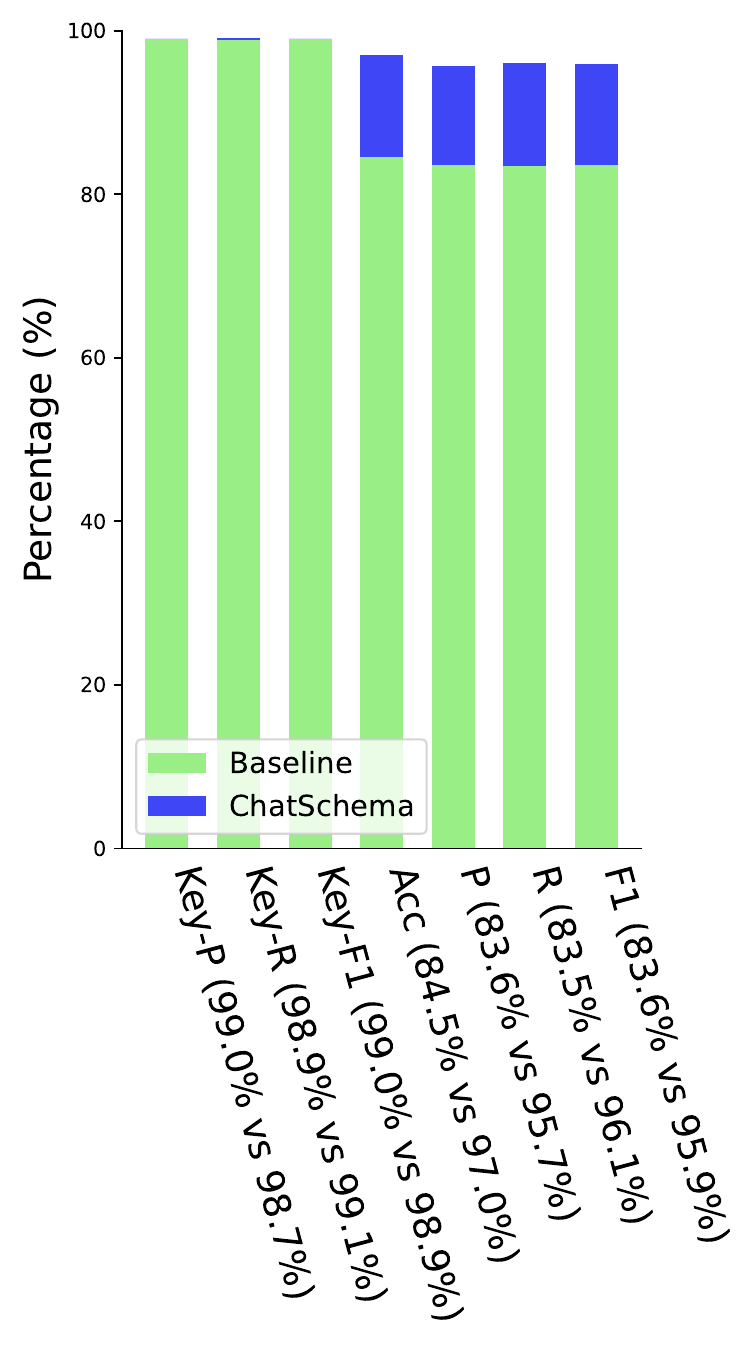}
    \caption{$Text$}
    \label{fig:gpt_text}
  \end{subfigure}
  \caption{Comparison of Baseline and ChatSchema with three different inputs on GPT-4o.}
  \label{fig:three graphs of gpt}
\end{figure}

\begin{figure}
  \centering
  \begin{subfigure}[b]{0.3\textwidth}
    \centering
    \includegraphics[width=\textwidth]{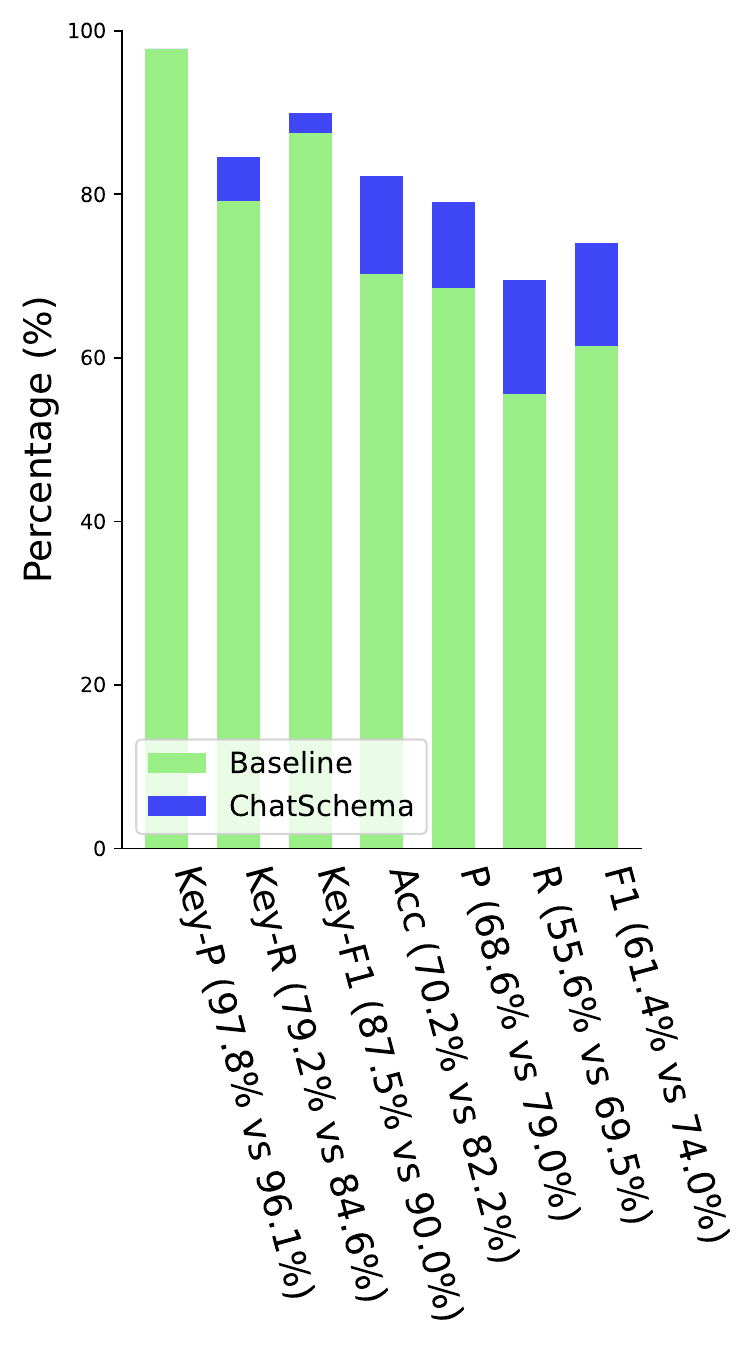}
    \caption{$Image$}
    \label{fig:gemini_image}
  \end{subfigure}
  \hfill
  \begin{subfigure}[b]{0.3\textwidth}
    \centering
    \includegraphics[width=\textwidth]{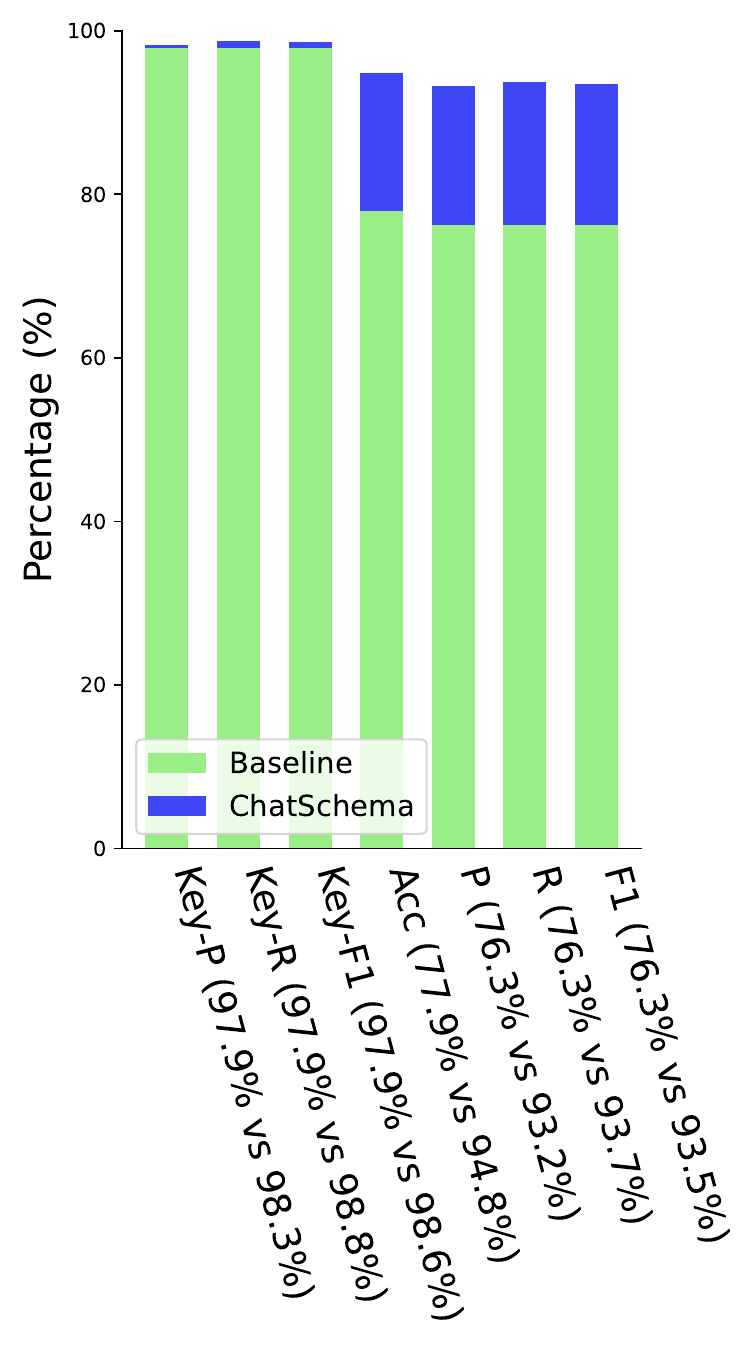}
    \caption{$Image + Text$}
    \label{fig:gemini_mixture}
  \end{subfigure}
  \hfill
  \begin{subfigure}[b]{0.3\textwidth}
    \centering
    \includegraphics[width=\textwidth]{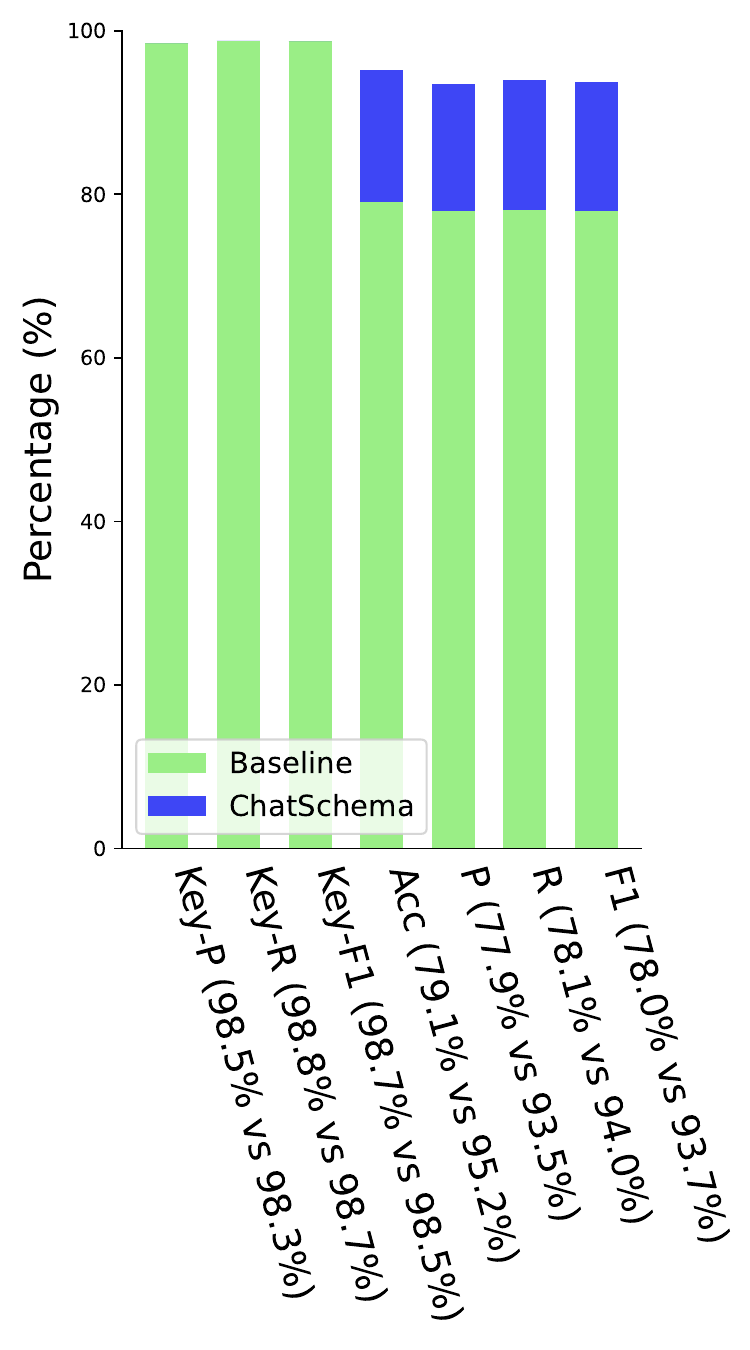}
    \caption{$Text$}
    \label{fig:gemini_text}
  \end{subfigure}
  \caption{Comparison of Baseline and ChatSchema with three different inputs on Gemini 1.5 Pro.}
  \label{fig:three graphs of gemini}
\end{figure}

The compare results between Baseline and ChatSchema on GPT and Gemini 1.5 Pro over three different input models are shown in \cref{fig:three graphs of gemini} and \cref{fig:three graphs of gpt}, respectively. Compared to Baseline, ChatSchema demonstrates significantly higher overall accuracy, precision, recall, and F1-score. Use the mixture of image and text on Gemini 1.5 Pro as an example, \cref{fig:gemini_mixture}: The performance metrics of key-precision, key-recall, key-F1-score for ChatSchema and Baseline are close, but ChatSchema still shows an edge. Moreover, ChatSchema achieves an overall accuracy of 94.8\% compared to Baseline's 77.9\%. The precision, recall, and F1-score for ChatSchema are  93.2\%, 93.7\%, and 76.3\% respectively, while Baseline achieves 76.3\%, 76.3\%, and 76.3\% respectively. This means that ChatSchema improves the Baseline with an increase of 26.9\% in the overall accuracy and 27.4\% in the overall F1-score, respectively.

\section{Discussion}
To identify the underlying errors and implement measures to mitigate them in future iterations, we analyze the Value mismatch, key-FP, and key-FN cases of our experiment result.

\begin{figure}
  \centering
  \includegraphics[width=0.8\linewidth]{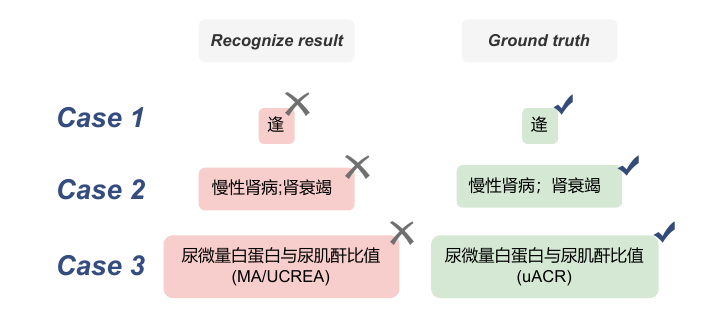}
  \caption{Example of error cases.}
  \label{fig:cases}
\end{figure}

For Value mismatch cases, the error stemmed from three main aspects. The first one is the report printing issues, for instance, some reports had dates printed as \enquote{2024-05-} with the latter part of the date missing or not printed clearly. This incomplete information often led the model to automatically fill in the missing parts incorrectly. The model's attempts to complete the date information resulted in mismatches between the ground truth and the predicted values. The second issue contributing to the Value mismatch cases is the challenge of recognizing rare Chinese characters, which are not commonly used in everyday language and thus are not well covered by the OCR module. An illustrative example of this issue is in \cref{fig:cases} case 1 the misrecognition of the surname, which is a rare character in Chinese, often being mistaken for the more common character by the OCR module. The third one is the OCR-related issues. For example, in the diagnosis of diseases, such as case 2 in \cref{fig:cases}, the Chinese semicolon was sometimes misrecognized as an English semicolon by the OCR module. These punctuation errors caused incorrect information extraction. The high occurrences indicate that OCR inaccuracies significantly impacted the model's performance. So the further direction to improve is the pre-correction module.

Besides, the main reason for key-FP and key-FN cases is the key match problem, which is due to the incorrect or absent association between specific medical terms and their corresponding identifiers in the schema. A case in point is case 3 in \cref{fig:cases}, the LMMs did not match the recognized result term with its equivalent key in the schema.

This study has several limitations. The dataset size is relatively small and sourced from only one hospital, leading to a lack of data diversity. The scope of report formats and content is limited. We could incorporate data from multiple hospitals and various types of medical documents in future work.

Our study demonstrates that the proposed method is highly effective in extracting structured information from medical reports. The method successfully performs tasks such as field name standardization, unit conversion, value mapping, and data type transformation. Through experiments, our results show that the method not only accurately extracts and processes information but also generalizes well across different LMMs, making it suitable for various base models.

The method provides a robust and efficient solution for extracting and structuring medical information from diverse medical scenarios, demonstrating significant potential for improving the utilization efficiency of medical reports. The method's robustness and adaptability to different models underline its possibility for widespread adoption in the medical field, offering a feasible solution for enhancing the effectiveness of medical data processing. Future research could further validate the method with more extensive and diverse datasets and explore its application in other domains and languages.

\begin{ack}
  This work was supported by the Ministry of Science and Technology of the People's Republic of China (2022YFF1203001).
\end{ack}

{\small
\bibliographystyle{unsrt}
\bibliography{references}
}

\newpage
\appendix

\section{Appendix}
\begin{figure}[hb]
  \centering
  \includegraphics[width=0.9\linewidth]{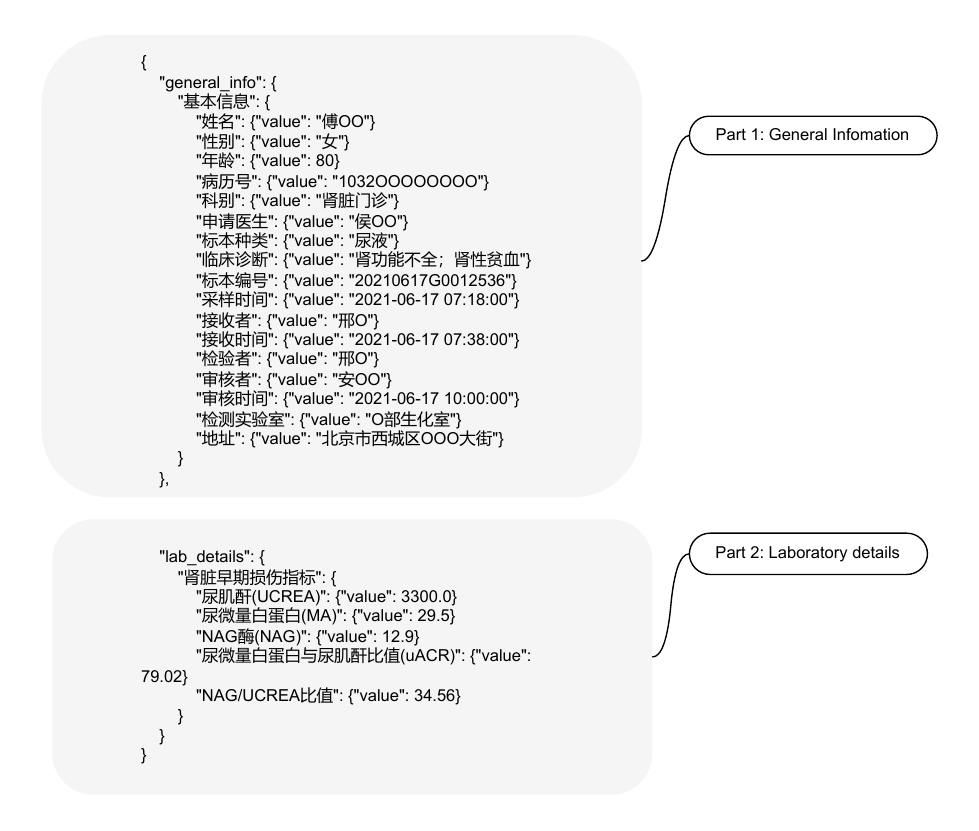}
  \caption{One case of the annotated dataset.}
  \label{fig:enter-label}
\end{figure}

\end{document}